\documentclass[letterpaper, 10pt, conference]{ieeeconf}

\IEEEoverridecommandlockouts
\usepackage{amsmath}
\usepackage{amssymb}
\usepackage{graphicx}
\usepackage{booktabs}
\usepackage{multirow}
\usepackage{xcolor}
\usepackage{hyperref}
\usepackage{pifont}
\usepackage{colortbl}

\newcommand{\cmark}{\ding{51}}
\newcommand{\xmark}{\ding{55}}

\definecolor{sevogreen}{RGB}{220,245,220}
\definecolor{basegray}{RGB}{235,235,235}

\title{\LARGE \bf
SEVO: Semantic-Enhanced Virtual Observation for Robust VLA Manipulation via Active Illumination and Data-Centric Collection\thanks{Code: \url{https://github.com/FelixFtch/SEVO}}
}

\author{Tianchonghui Fang$^{1}$, Yuan Zhuang$^{1}$, and Fei Miao$^{1*}$
\thanks{$^{1}$T.~Fang, Y.~Zhuang, and F.~Miao are with the School of Computing, University of Connecticut, Storrs, CT 06268, USA. Email: {\tt\small ftchftch@gmail.com, yuan.2.zhuang@uconn.edu, miaofeicat@hotmail.com}.}
\thanks{$^{*}$Corresponding author: Fei Miao (\texttt{miaofeicat@hotmail.com}).}
}

\begin{document}

\maketitle
\thispagestyle{empty}
\pagestyle{empty}

\begin{abstract}

Vision-Language-Action (VLA) and imitation-learning policies trained via community toolchains on low-cost hardware frequently fail when deployed outside the training environment. Existing evaluations, including the original ACT and SmolVLA benchmarks, typically demonstrate high success rates under controlled, fixed backgrounds with high-contrast objects, yet community practitioners consistently report near-zero transfer to new environments. We present \textbf{SEVO} (Semantic-Enhanced Virtual Observation), a data-centric approach that substantially improves cross-environment manipulation robustness \textit{without modifying the policy architecture}. SEVO transforms the raw RGB camera stream through three synergistic mechanisms: (1)~body-fixed cameras whose combined fields of view cover the full manipulation workspace, (2)~active red-spectrum illumination that physically normalizes object appearance, and (3)~real-time YOLO segmentation overlay that provides a background-invariant semantic cue. Critically, we show that a diversified data collection protocol (systematically varying lighting, backgrounds, and distractors during teleoperation) is the single most important factor for generalization. We deliberately target transparent water bottles, objects that visually blend with their surroundings, and select a simple pick-and-place task to enable hundreds of controlled real-robot trials across two mobile platforms. The full pipeline achieves 95\% grasp success with ACT and 83\% with SmolVLA in the training environment, transferring to visually distinct novel environments at 85\% and 75\%. Without SEVO, the same policies achieve only 75\%/70\% in training and collapse to 30--35\% in novel environments, with zero transfer to extreme settings. Our results demonstrate that principled observation design and environmental diversity during data collection, not model scaling, enable low-cost robots to operate reliably in everyday household environments.

\end{abstract}

\section{INTRODUCTION}

Community-driven frameworks like LeRobot~\cite{lerobot2024} with low-cost robotic arms (e.g., SO-101) have democratized imitation learning, enabling small labs and individuals to train policies such as ACT~\cite{zhao2023act} and SmolVLA~\cite{shukor2025smolvla}. However, a persistent frustration reported is that trained policies only work in the exact environment where the data was collected. Changes in the background, lighting, or camera viewpoint can cause deployment failure. Benchmarks indicate that ACT trained on 50 episodes with a fixed background achieves about 50--70\% success rate on single-task pick-and-place, with near-zero success rate when transfering to new environments~\cite{lerobot2024}. SmolVLA's official evaluation reports success rate of 78.3\% for SmolVLA and 48.3\% for ACT on a multi-task SO-100 benchmark~\cite{shukor2025smolvla}, but these numbers still assume matched train/test environments.

This fragility is not unique to community setups like LeRobot. The original ACT evaluation~\cite{zhao2023act} places objects along a ``white reference line'' on a clean tabletop, with a tarp blocking background distractions. InterACT~\cite{lee2024interact} similarly uses ``a tarp around the setup to block unnecessary background.'' SmolVLA's real-world evaluation~\cite{shukor2025smolvla} tests on fixed SO-100/SO-101 setups. Literature about generalization gap~\cite{hsu2023generalization} systematically shows that background changes cause significant degradation in imitation-learning policies. Causal-ACT~\cite{causalact2025} explicitly identifies background as a source of ``causal confusion'' where the policy overfits to task-irrelevant surroundings. In summary, reported success rates largely depend on the controlled backgrounds with high-contrast objects (e.g., colored blocks on white surfaces), conditions far from real-world deployment.

To demonstrate that this background-generalization problem can be solved at the \textit{observation level}, we deliberately chose transparent water bottles as our manipulation target. Their specular, reflective surfaces make the task particularly challenging, since they visually blend with their surroundings. Rather than selecting high-contrast tasks, we target the harder scenario where object--background separation requires active visual engineering. We pair this with a simple pick-and-place task that enables rapid, large-scale comparison, experimentation, and ablation study, based on hundreds of real-robot trials across systematically varied environments.

Our approach, SEVO (Semantic-Enhanced Virtual Observation), redesigns the observation pipeline and data collection protocol while keeping the downstream policy model architecture \textit{unchanged}. The key insight is threefold: (1)~cameras mounted at positions that are \textbf{static relative to the robot body} provide stable spatial context; (2)~\textbf{active red illumination} physically stabilizes the appearance of transparent objects; (3)~our proposed \textbf{real-time YOLO segmentation overlay} creates background-invariant semantic cues. Most importantly, our ablation shows that a \textbf{diversified data collection protocol} (including varying backgrounds, lighting, and distractors) is the single most important factor, outranking both illumination and overlay.

We validate SEVO on two physically distinct mobile platforms, achieving success rates of 95\% (ACT) and 83\% (SmolVLA) in the training environment, with graceful degradation to 85\% and 75\% in unseen environments, compared to only 30--35\% for the same policies without SEVO. Through parameter-level inspection, we explain \textit{why} these observation-space modifications benefit ACT's fully trainable backbone more than SmolVLA's frozen vision encoder.

In summary, our contributions are:
\begin{enumerate}
    \item We propose a \textbf{data-centric observation pipeline} that integrates YOLO overlay, active red illumination, and body-fixed cameras, to improve generalization of low-cost robots across visually diverse environments, with \textbf{parameter-level analysis} explaining how these modifications propagate through trainable vs.\ frozen policy architectures.
    \item We design a \textbf{diversified data collection protocol} and show that it is the dominant factor in cross-environment generalization, validated across two robots and two policies with hundreds of real-robot trials.
    \item Through experiments, we provide systematic evidence that \textbf{body-fixed cameras substantially outperform the community-default wrist cameras}, which induce severe policy-dependent failures with distinct mechanisms for ACT and SmolVLA.
\end{enumerate}

\section{RELATED WORK}

\textbf{VLA and ACT for Manipulation.}
RT-2~\cite{brohan2023rt2} demonstrated that large vision-language models can directly generate robot actions. SmolVLA~\cite{shukor2025smolvla} targets affordable robotics with a compact VLA. ACT~\cite{zhao2023act} addresses compounding errors through action chunking. Octo~\cite{octo2024} achieves generalization via multi-robot pretraining. Despite progress, deployment remains fragile to visual domain shifts, and existing evaluations predominantly use fixed, clean backgrounds~\cite{zhao2023act,shukor2025smolvla,lee2024interact}. We address this fragility from the perception side without modifying any policy architecture.

\textbf{Environment Generalization in Imitation Learning.}
Hsu et al.~\cite{hsu2023generalization} decompose the generalization gap and show that background and table-texture shifts cause the largest performance drops. Causal-ACT~\cite{causalact2025} identifies ``causal confusion'' where policies rely on task-irrelevant background features. Domain randomization~\cite{tobin2017domain} and sim-to-real~\cite{yang2024harmonic} approaches improve generalization through training-side variation but require simulation infrastructure. SEVO pursues a similar goal through a purely data-centric approach on real hardware, avoiding the sim-to-real gap entirely. This is particularly important for transparent objects, whose specular reflections and detector noise are difficult to model accurately in simulation.

\textbf{The Colosseum.}
Pumacay et al.~\cite{pumacay2024colosseum} introduce a benchmark that evaluates manipulation generalization under systematic environment perturbations (including appearance and scene changes), and report substantial performance degradation across shifts. This reinforces our motivation that matched train/test scenes can overstate robustness.

\textbf{Data Scaling Laws.}
Lin et al.~\cite{lin2024data_scaling} study scaling behavior in imitation learning and show that \emph{environment diversity} is a key driver of transfer. This directly supports our central experimental finding that varied backgrounds are the dominant factor in SEVO's generalization gains.

\textbf{GreenAug.}
Teoh et al.~\cite{teoh2024greenscreen} propose green-screen-based data collection with chroma-key background replacement to improve scene generalization. In contrast, SEVO keeps the training and deployment environments fully real and targets difficult, lighting-sensitive transparent objects, where physical reflections and detector noise are hard to reproduce with purely synthetic edits.

\textbf{RoboEngine.}
Yuan et al.~\cite{yuan2025roboengine} augment robot data via semantic robot segmentation and background generation, enabling plug-and-play scene variation from limited real demonstrations. This is complementary to SEVO: while RoboEngine expands backgrounds through generation, SEVO emphasizes real-world collection under diverse scenes and uses active illumination plus online observation transformation to stabilize semantic cues under complex lighting that is difficult to model in simulation.

\textbf{Visual Prompting and Intermediate Representations.}
MOO~\cite{stone2023moo} interfaces policies with VLMs for object specification. MOKA~\cite{liu2024moka} proposes mark-based visual prompting. OmniVLA~\cite{guo2025omnivla} overlays sensor-derived masks onto RGB. Our work shares the overlay philosophy but targets low-cost hardware, combines detector-driven masks with physical spectral conditioning, and emphasizes data protocol as the primary lever.

\textbf{Policy Baselines and Edge Deployment.}
Large-scale VLAs such as $\pi_{0.5}$~\cite{pi05_2025} and OpenVLA~\cite{kim2024openvla} require billions of parameters and cannot run on edge hardware. We evaluate on ACT (51.60M, fully trainable) and SmolVLA (450.05M, 22.2\% trainable), the two dominant architectures deployable on community-accessible hardware (Jetson Orin NX, Raspberry Pi~5), trainable from 80--120 episodes.

\begin{figure*}[t]
    \centering
    \includegraphics[width=\textwidth]{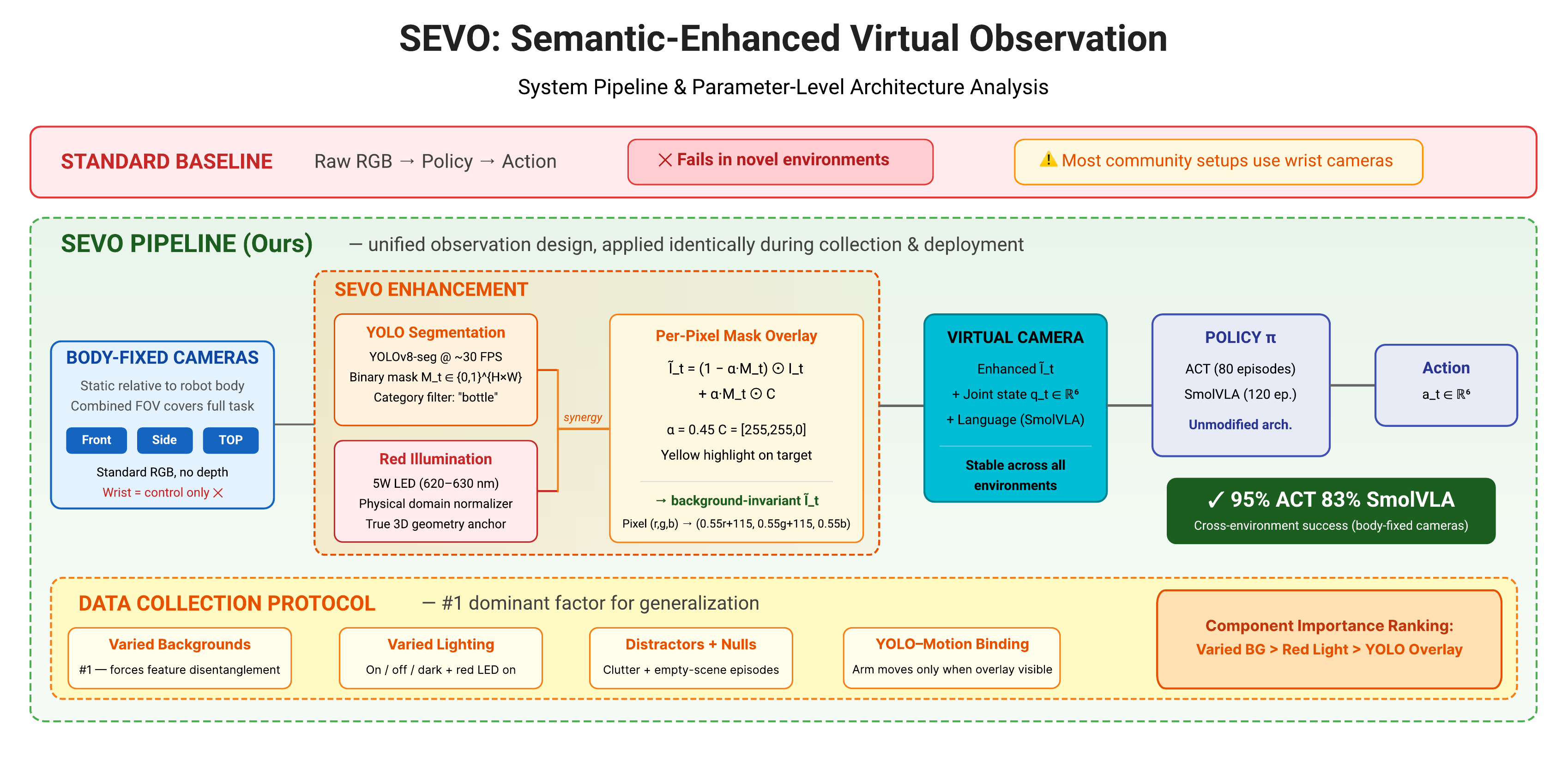}
    \caption{\textbf{SEVO system overview.} \textbf{Top:} Standard baseline (raw RGB $\to$ policy) fails in novel environments. \textbf{Middle:} The SEVO pipeline transforms raw RGB from body-fixed cameras into a background-invariant virtual camera stream $\tilde{I}_t$ via three mechanisms: YOLOv8-seg mask overlay ($\alpha{=}0.45$, yellow), 5\,W red LED illumination (620--630\,nm), and a diversified data collection protocol. \textbf{Bottom-left:} ACT (51.60M, 100\% trainable). SEVO-modified pixels directly contact the trainable \texttt{conv1} $[64,3,7,7]$, adapting ImageNet-pretrained color filters end-to-end. \textbf{Bottom-right:} SmolVLA (450.05M, 22.2\% trainable). SEVO pixels enter frozen SigLIP $[768,3,16,16]$ patch embedding, compressing each $16{\times}16$ patch into one token; only the Action Expert (98.2M) is trainable. \textbf{Bottom:} Robot~A (Jetson Orin NX, 2 body-fixed cameras) and Robot~B (Raspberry Pi~5, 3 body-fixed cameras).}
    \label{fig:overview}
\end{figure*}

\section{PLATFORMS AND TASK}

\subsection{Robot Platforms}

\begin{figure}[t]
    \centering
    \includegraphics[width=\columnwidth]{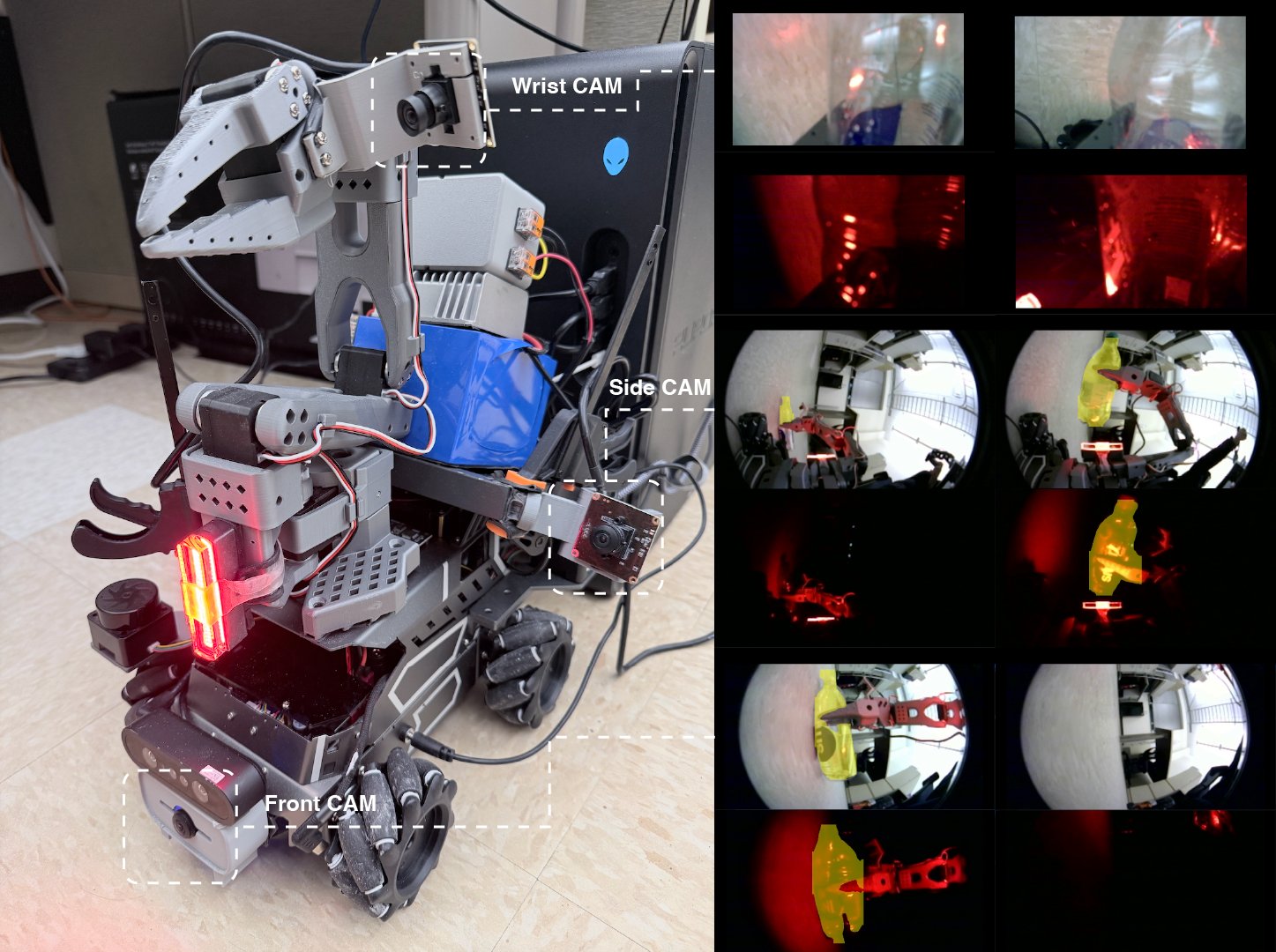}
    \caption{\textbf{Robot~A:} primary evaluation platform. \textbf{Left:} The mobile chassis navigates to diverse locations via LiDAR-based path planning to evaluate SEVO-trained policies in different environments. Two body-fixed cameras (Front~CAM, Side~CAM) serve as policy inputs; a separate black detection camera (front, visible next to the front cam) is used only by the chassis controller to detect bottles and trigger a stop. Red LED visible at arm base. Only the SO-101 arm is trained; the chassis is controlled by a separate path-planning module. During evaluation, we follow the robot and manually place bottles in front of it at each location. \textbf{Right:} Camera views showing wrist camera views (top rows, information-poor at close range, included only for ablation), side camera with YOLO overlay (middle rows, yellow highlight visible in both ambient and dark conditions), and front camera views (bottom rows).}
    \label{fig:robotA}
\end{figure}

\begin{figure}[t]
    \centering
    \includegraphics[width=\columnwidth]{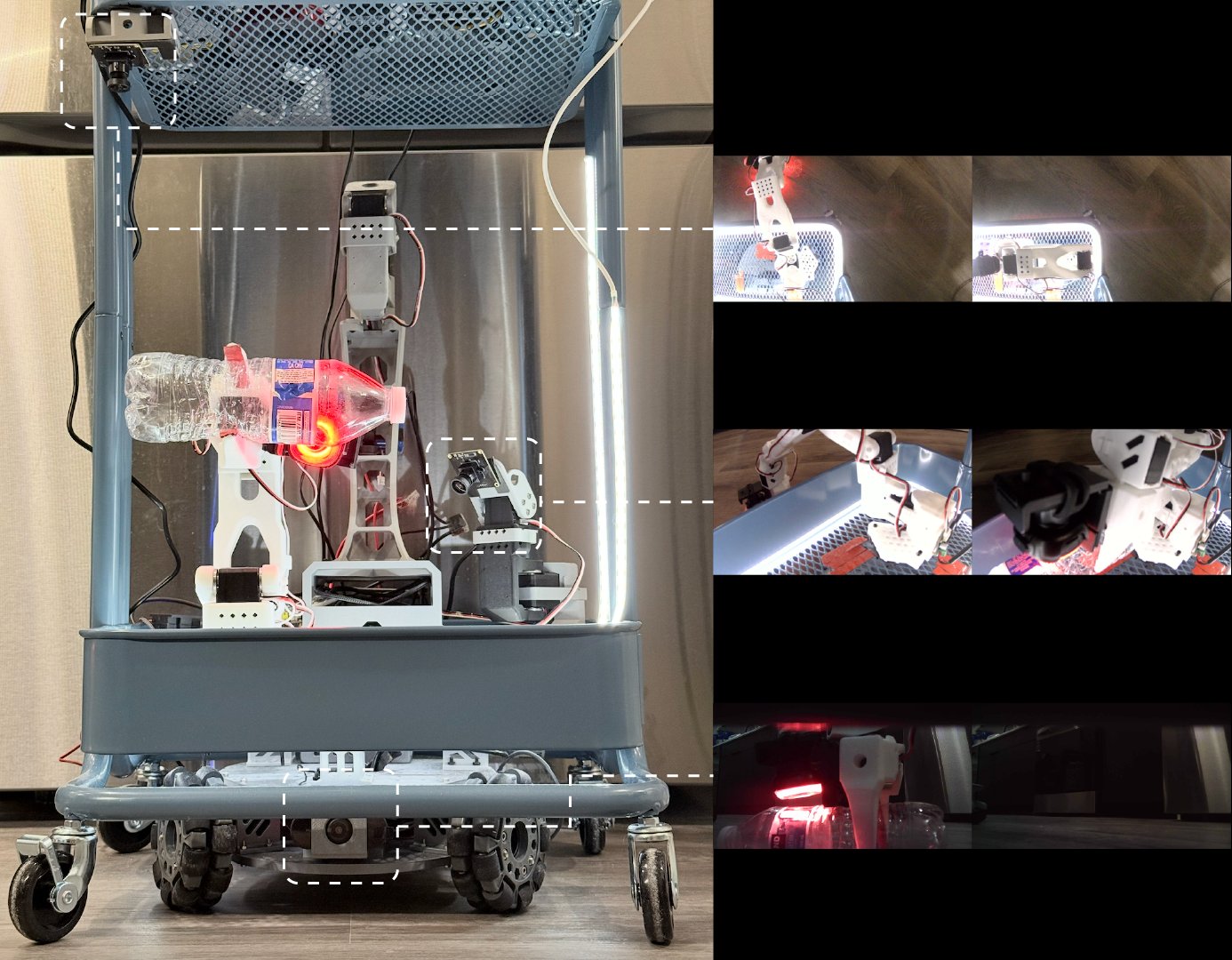}
    \caption{\textbf{Robot~B:} cross-platform validation. This platform uses a \textit{completely different} camera arrangement (3 body-fixed: overhead, side, front) and hardware design (IKEA cart, Raspberry Pi~5) from Robot~A, yet is trained with the same SEVO protocol. Despite extremely slow inference (ACT: ${\sim}$2\,min, SmolVLA: ${\sim}$15\,min per grasp), Robot~B achieves comparable success rates in cluttered, varied-background environments. \textbf{Left:} Three camera positions with overhead red LED creating gradient illumination. \textbf{Right:} Camera views from each position: basket interior (overhead), arm workspace (side), front approach (front). Robot~B confirms that SEVO's effectiveness is not coincidental to Robot~A's specific setup, but transfers across robot morphologies, camera layouts, and compute platforms.}
    \label{fig:robotB}
\end{figure}

We validate SEVO on two physically distinct mobile manipulation platforms (Fig.~\ref{fig:overview}, \ref{fig:robotA}, \ref{fig:robotB}), both equipped with SO-101 arms and using the LeRobot~\cite{lerobot2024} software stack.

\textbf{Robot~A} (primary evaluation platform): Custom autonomous car with Jetson Orin NX (16\,GB). Two body-fixed cameras (front, side) plus a 5\,W red LED (620--630\,nm) at the arm base. Runs ACT in real time; SmolVLA requires ${\sim}$2\,min per cycle. A separate front-facing detection camera is used only by the chassis path-planning module to trigger a stop when a bottle is detected ahead. The arm policy itself is the only component trained via SEVO.

\textbf{Robot~B} (cross-platform validation): Redesigned IKEA LeRobot cart with Raspberry Pi~5 (8\,GB). Three body-fixed cameras (front, side, overhead) in a completely different arrangement from Robot~A, plus an overhead red LED providing distance-dependent gradient illumination. Due to compute limits, ACT requires ${\sim}$2\,min and SmolVLA ${\sim}$15\,min per grasp cycle.

All cameras are standard RGB ($640{\times}480$), body-fixed (static relative to the robot), and their combined fields of view cover the full manipulation sequence. YOLOv8-seg~\cite{jocher2023yolo} runs at ${\sim}$30\,FPS on Robot~A.

\subsection{Camera Placement Principle}

Unlike most community setups that default to wrist-mounted cameras, we require all cameras to be \textbf{mounted at positions static relative to the robot body}. Individual cameras need not capture the full workspace; rather, their \textbf{combined fields of view must cover the entire manipulation sequence}, from initial approach through grasping to final placement. This ensures that: (1)~the visual observation does not change as a function of arm state, eliminating a major source of distributional shift; and (2)~each camera captures a consistent spatial relationship between the gripper, target object, and placement goal.

\subsection{Task Definition}

We deliberately select a \textbf{simple, repeatable pick-and-place task} to enable rapid, large-scale controlled experimentation: detect a transparent water bottle in front of the robot, stop the mobile base for ${\sim}$2\,s, grasp the bottle, and place it to the side (Robot~A) or into a basket (Robot~B). We chose this simple task because it enables hundreds of real-robot trials with systematic variation of a single factor (visual environment), isolating SEVO's observation-space effect while keeping arm kinematics nearly identical across all trials. The choice of a transparent water bottle, rather than a high-contrast colored block, ensures that the task is genuinely challenging: the bottle visually blends with most backgrounds, making it a stringent test of background-invariant grasping.

\section{METHOD: SEVO}

SEVO transforms raw RGB into a virtual camera output with stable semantic cues, applied identically during data collection and deployment. It operates \textit{exclusively in observation space}, modifying what the policy sees, not how it learns.

\subsection{Pipeline Overview}

The pipeline (Fig.~\ref{fig:overview}) processes each body-fixed camera feed:
\begin{equation}
    I_t \xrightarrow{\text{YOLO}} M_t \xrightarrow{\text{overlay}} \tilde{I}_t \;\;\Rightarrow\;\; o_t = \{\tilde{I}_t,\; q_t\} \xrightarrow{\pi} a_t
    \label{eq:pipeline}
\end{equation}
where $I_t \in \mathbb{R}^{H \times W \times 3}$ is the raw RGB frame, $M_t \in \{0,1\}^{H \times W}$ is the YOLO segmentation mask for ``bottle,'' $\tilde{I}_t$ is the enhanced image, $q_t \in \mathbb{R}^{6}$ is the 6-DOF joint state, $\pi$ is the unmodified policy (ACT or SmolVLA), and $a_t \in \mathbb{R}^{6}$ is the output action. The raw input $I_t$ is modified only in pixel space; the joint state $q_t$ passes through unchanged; and the policy $\pi$ receives \textit{exactly the same input format} as standard training; no policy model architectural modification is needed.

\subsection{Real-Time Semantic Overlay}

The enhanced image $\tilde{I}_t$ is computed per-pixel as:
\begin{equation}
    \tilde{I}_t = (1 - \alpha \cdot M_t) \odot I_t + \alpha \cdot M_t \odot C
    \label{eq:overlay}
\end{equation}
where $\alpha{=}0.45$ is the overlay opacity, $\odot$ is element-wise multiplication, and $C{=}[255,255,0]$ (yellow) is the highlight color. Within the detected region ($M_t{=}1$), each pixel $(r,g,b)$ becomes $(0.55r{+}114.75,\; 0.55g{+}114.75,\; 0.55b)$---a distinctive high-R, high-G, low-B signature largely independent of the background.

\textbf{YOLO--motion binding.} During data collection, the operator initiates arm motion \textit{only when the yellow overlay is visible}, creating training data where overlay presence is a necessary precondition for arm activation. At deployment, this learned binding acts as a safety gate: the arm activates only when YOLO detects the target. We also observe that policies trained with YOLO overlay successfully grasp bottles of different shapes and brands not seen during training, indicating that the policy anchors on the yellow mask rather than specific object appearance.

\subsection{Active Red-Spectrum Illumination}

A 5\,W LED (620--630\,nm) provides physical domain normalization. The LED produces consistent specular reflections on transparent surfaces that persist across backgrounds, improving YOLO detection. Its short-range spectral dominance suppresses ambient lighting variation, while the reflections serve as a physical visual anchor that encodes true 3D surface geometry. On Robot~A, the LED illuminates from the arm base; on Robot~B, an overhead LED creates distance-dependent gradient illumination that acts as an implicit proximity cue.

\subsection{How SEVO Propagates Through Policy Architectures}
\label{sec:architecture}

Both policies receive SEVO-modified pixels as standard RGB input (Table~\ref{tab:arch}), but the pixels encounter fundamentally different computational pathways.

\begin{table}[t]
\caption{Policy Architecture Comparison (Measured)}
\label{tab:arch}
\centering
\setlength{\tabcolsep}{3pt}
\footnotesize
\begin{tabular}{lrrl}
\toprule
\textbf{Component} & \textbf{Params} & \textbf{\%} & \textbf{Status} \\
\midrule
\multicolumn{4}{l}{\textit{ACT (51.60M total, 100\% trainable)}} \\
\;\; ResNet-18 backbone (shared) & 11.18M & 21.7 & \cmark Trainable \\
\;\; CVAE encoder (4-layer Trans.) & 13.28M & 25.7 & \cmark Trainable \\
\;\; Transformer encoder (4-layer) & 13.28M & 25.7 & \cmark Trainable \\
\;\; Transformer decoder (1-layer) & 5.91M & 11.5 & \cmark Trainable \\
\;\; Projections + action head & 0.34M & 0.7 & \cmark Trainable \\
\midrule
\multicolumn{4}{l}{\textit{SmolVLA (450.05M total, 22.2\% trainable)}} \\
\;\; SigLIP vision encoder (12-layer) & 85.1M & 18.9 & \xmark Frozen \\
\;\; SigLIP$\to$LM connector & 11.8M & 2.6 & \xmark Frozen \\
\;\; SmolLM2 decoder (16-layer) & 251.6M & 55.9 & \xmark Frozen \\
\;\; Action Expert (16-layer Trans.) & 98.2M & 21.8 & \cmark Trainable \\
\;\; State/action proj.\ + time MLPs & 1.64M & 0.4 & \cmark Trainable \\
\bottomrule
\end{tabular}
\vspace{1mm}

\footnotesize{Parameter counts from live model inspection.}
\end{table}

\textbf{ACT: direct pixel-to-weight pathway.} All 51.60M parameters are trainable. SEVO-modified pixels enter \texttt{conv1} layer, the first convolutional layer of the ResNet-18 backbone, whose ImageNet-pretrained weights encode color-sensitive features. During fine-tuning on SEVO data, these filters adapt end-to-end: the overlay's high-R, high-G, low-B signature and the red LED's specular highlights provide consistent gradient signal that reshapes \texttt{conv1} into SEVO-specific color detectors. The Transformer encoder/decoder (26.56M) maps visual features to chunked actions through a fully differentiable pathway from pixel to action.

\textbf{SmolVLA: frozen encoder bottleneck.} SEVO pixels enter the frozen SigLIP vision encoder, which compresses each image patch into a fixed-length token. Since this encoder was pretrained on web image and text data rather than robotics observations, it was never optimized for SEVO's overlay pattern. The resulting tokens pass through the entire frozen VLM backbone before reaching the trainable Action Expert. The Expert must therefore decode SEVO cues from compressed, indirect features, explaining both the higher data requirement (120 vs. 80 episodes) and the lower ceiling performance (83\% vs. 95\%).

\section{DATA COLLECTION PROTOCOL}

\subsection{Demonstration Strategy}

We collect teleoperated episodes via leader--follower control: 80 for ACT and 120 for SmolVLA. The robot remains at a fixed position during data collection in a single domestic room with dark-toned flooring. The arm trajectory remains consistent (bottle at fixed position, same grasp motion) while we systematically vary the visual environment:

\textbf{Backgrounds:} dark/bright backdrops, different floors, cluttered/clean surroundings. This is the dominant factor (Section~\ref{sec:component_ranking}).

\textbf{Lighting:} room lights on/off, completely dark with only red LED active, varied ambient directions.

\textbf{Distractors and human presence:} random objects placed around the robot; people walking through the scene during recording. These explicitly teach the policy to ignore non-target visual changes.

\textbf{Null episodes:} scenes with no bottle, where the arm remains stationary. Combined with the YOLO--motion binding (the operator waits ${\sim}$1\,s after the yellow overlay appears before initiating arm motion), this teaches the policy to activate \textit{only} upon confirmed bottle detection.

\textbf{Object pose:} bottle angle and position vary slightly within the reachable workspace; the primary variation emphasis is on environment, not object placement.
The goal of this protocol is to ensure that the only consistent visual element across episodes is the bottle and its SEVO-enhanced appearance. Background objects, lighting, and bystanders all change between episodes, teaching the policy that these elements are task-irrelevant. This kind of within-scene visual diversity is difficult to replicate faithfully in simulation, where rendered clutter, human motion, and lighting lack photorealistic variation.

\subsection{Training Details}

We train both policies using the default LeRobot~\cite{lerobot2024} recipes without architectural modifications (no LoRA or adapters). ACT~\cite{zhao2023act} is trained on 80 episodes for 400K steps with batch size 24, using the AdamW optimizer with default learning rate and weight decay disabled, and mixed-precision training enabled. SmolVLA~\cite{shukor2025smolvla} is trained in 120 episodes for 20K steps with batch size 64, fine-tuned from the official pretrained checkpoint. Language prompt: "pick up the bottle and place it on the side." All training is performed on a single NVIDIA RTX 4090. Both policies receive SEVO-enhanced $\tilde{I}_t$ from all body-fixed cameras and joint state $q_t$.

\section{EXPERIMENTS}
\label{sec:exp}

\subsection{Motivation and Experimental Design}

The VLA and LeRobot communities consistently report that even well-performing policies fail when the visual environment changes~\cite{lerobot2024,causalact2025,hsu2023generalization}. The original ACT paper~\cite{zhao2023act} evaluates on a clean tabletop with tarps blocking background distractions; SmolVLA~\cite{shukor2025smolvla} evaluates on fixed SO-100/SO-101 setups; and InterACT~\cite{lee2024interact} explicitly uses background-blocking tarps. Many published evaluations use high-contrast objects (e.g., colored cubes or blocks on white surfaces~\cite{zhao2023act,shukor2025smolvla}) that can be trivially segmented, ignoring the distribution shift challenges of household deployment.

Our experiments evaluate whether SEVO can close the background-generalization gap identified above. We use transparent water bottles as the target object because they blend with most backgrounds, and a pick-and-place task that allows hundreds of controlled real-robot trials to isolate each variable. We report success rates over 100 trials per condition.

We define three evaluation settings based on visual similarity to the training room:

\begin{itemize}
    \item \textbf{Training environment:} the same domestic room where data was collected (dark flooring, familiar furniture). The robot may navigate to different viewpoints within this room via chassis path planning, but the room itself is unchanged.
    \item \textbf{Novel (similar):} rooms the robot has never visited during data collection, with different furniture, wall colors, lighting fixtures, and clutter, but comparable dark-toned flooring. Test locations include a bathroom, apartment hallway, university corridor (gray wood floor), classroom (patterned dark carpet), and school building lobby (dark green tile).
    \item \textbf{Novel (extreme):} environments whose floor tone falls entirely outside the training distribution. In our case, a laboratory with bright white flooring, which was absent from all training episodes.
\end{itemize}

The shared property of all ``similar'' environments is dark-toned flooring, which matches the training room. The ``extreme'' label reflects a specific out-of-distribution factor (bright floor tone) rather than overall visual complexity. We note that when we separately collected 80 SEVO episodes \textit{in} the white-floor laboratory and trained a new policy, the robot grasped successfully in that setting, suggesting that including bright-floor scenes in the original training distribution would likely close this gap.

At deployment with Robot~A's path-planning chassis, the robot navigates to diverse locations in these unseen environments. We observe that: the arm remains stationary during navigation (no false triggers), does not activate when clutter or human hands appear (no misfire), and grasps the bottle with high success regardless of whether lights are on or off, activating \textit{only} when a bottle is detected in front. Robot~B independently validates these results with a completely different hardware design, confirming that SEVO generalizes across robot platforms.

\subsection{Community Baselines}

ACT on 50 episodes with fixed backgrounds achieves success rate of ${\sim}$50--70\% on single-task pick-and-place~\cite{lerobot2024}. SmolVLA's official benchmark reports 78.3\% (SmolVLA) and 48.3\% (ACT from scratch)~\cite{shukor2025smolvla}. Our no-SEVO baseline (Table~\ref{tab:ablation}, bottom row) of 63\% (ACT) and 51\% (SmolVLA) is consistent with these community reports, given our harder target task (transparent bottle).

\subsection{Component Ablation Study}

Table~\ref{tab:ablation} presents the ablation study of SEVO components using body-fixed cameras only. Each cell averages training-environment and similar-novel-environment trials.

\begin{table}[t]
\caption{Component Ablation: Body-Fixed Cameras Only}
\label{tab:ablation}
\centering
\setlength{\tabcolsep}{3.5pt}
\begin{tabular}{ccccc}
\toprule
\textbf{YOLO} & \textbf{Red Light} & \textbf{Varied BG} & \textbf{ACT (\%)} & \textbf{SmolVLA (\%)} \\
\midrule
\rowcolor{sevogreen}
\cmark & \cmark & \cmark & \textbf{90}$^{\dagger}$ & \textbf{79}$^{\dagger}$ \\
\xmark & \cmark & \cmark & 87 & 76 \\
\cmark & \xmark & \cmark & 84 & 71 \\
\cmark & \xmark & \xmark & 75 & 61 \\
\xmark & \cmark & \xmark & 66 & 53 \\
\rowcolor{basegray}
\xmark & \xmark & \xmark & 63$^{*}$ & 51$^{*}$ \\
\bottomrule
\end{tabular}
\vspace{1mm}

\vspace{1mm}
\begin{minipage}{\columnwidth}
\footnotesize
ACT: 80 training episodes; SmolVLA: 120 training episodes. $N{=}100$ trials per condition on Robot~A, averaged across training and similar novel environments.

$^{\dagger}$\textbf{Full SEVO}: the complete pipeline combining all three components.

$^{*}$\textbf{Baseline}: standard RGB input without any SEVO enhancement.

Full SEVO raises ACT from 63\% to 90\% and SmolVLA from 51\% to 79\%.
\end{minipage}
\end{table}

\textbf{Key findings.} Full SEVO achieves higher success rate across various environments than unmodified ACT/SmolVLA. Diversified backgrounds in the training data are the dominant factor: removing them costs 9--21 points for ACT and 10--23 points for SmolVLA. Red light contributes 3--6 points. YOLO overlay adds 3 points numerically, but provides categorical filtering that restricts grasping to detected bottles; without it, policies also grasp other reflective objects. All SEVO conditions achieve zero false triggers on empty scenes.

\subsection{Component Importance Ranking}
\label{sec:component_ranking}
Through experiments, we observe that each module of SEVO contributes to the enhanced generalization capability on the hardware, and the effectiveness can be ranked as
\begin{equation}
    \underbrace{\text{Varied BG}}_{\text{dominant}} \;>\; \underbrace{\text{Red Light}}_{\text{second}} \;>\; \underbrace{\text{YOLO Overlay}}_{\text{third}} \nonumber
\end{equation}
This ordering is consistent across both policies and both robot platforms. SmolVLA degrades more than ACT when any component is removed, which we attribute to the frozen-encoder bottleneck described in Section~\ref{sec:architecture}.

\subsection{Cross-Environment Transfer}

\begin{table}[t]
\caption{Cross-Environment Transfer (Full SEVO vs.\ No SEVO)}
\label{tab:transfer}
\centering
\setlength{\tabcolsep}{3.5pt}
\begin{tabular}{lccc}
\toprule
\textbf{Policy} & \textbf{Train Env.} & \textbf{Novel (similar)} & \textbf{Novel (extreme)} \\
\midrule
\rowcolor{sevogreen}
ACT w/ SEVO     & \textbf{95} & \textbf{85} & 10 \\
\rowcolor{sevogreen}
SmolVLA w/ SEVO & \textbf{83} & \textbf{75} & 5 \\
\midrule
ACT (no SEVO)     & 75 & 30 & 0 \\
SmolVLA (no SEVO) & 70 & 35 & 0 \\
\bottomrule
\end{tabular}
\vspace{1mm}

\vspace{1mm}
\begin{minipage}{\columnwidth}
\footnotesize
Novel (similar): unseen rooms with different furniture, lighting, and clutter but comparable floor tones.

Novel (extreme): drastically different environments (e.g., pure white floor, highly reflective surfaces).

Without SEVO, both policies drop to 30--35\% in novel environments and 0\% in extreme settings. With SEVO, novel-environment success remains at 85\%/75\%. The gap in extreme environments reflects out-of-distribution floor tones absent from the training data.
\end{minipage}
\end{table}

\begin{figure}[t]
    \centering
    \includegraphics[width=\columnwidth]{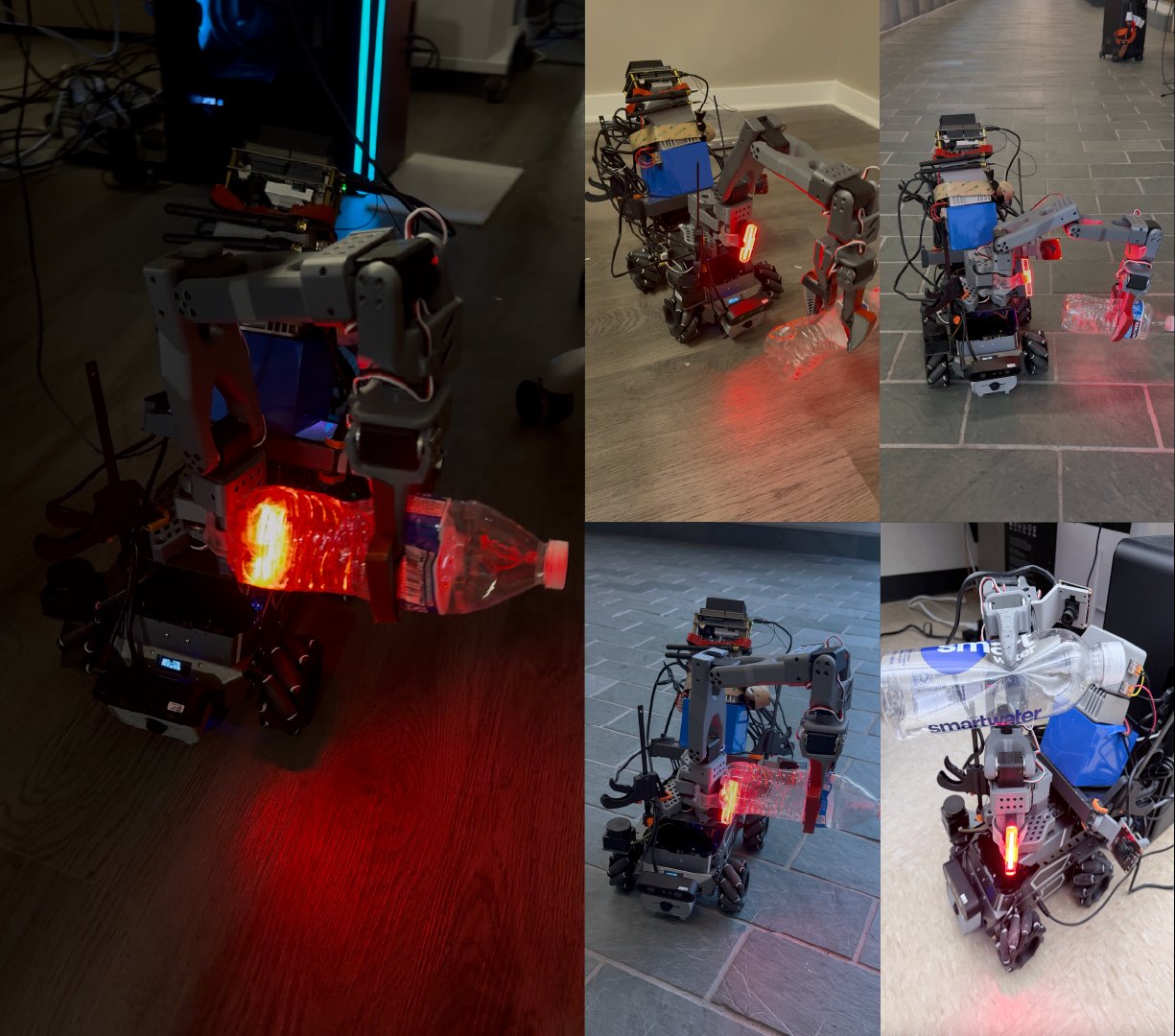}
    \caption{Cross-environment deployment of Robot~A with full SEVO (corresponding to the SEVO rows in Table~\ref{tab:transfer}). \textbf{Left to right:} training environment and progressively different unseen environments with varied backgrounds, lighting, floors, and clutter. The robot successfully grasps water bottles across all settings. \textbf{Bottom-right:} an extreme case with drastically different white floor where success rate drops sharply. We also observed that with SEVO, the policy generalizes to \textit{other brands and shapes of bottles} highlighted by YOLO, which confirms the policy anchors on the semantic overlay rather than specific object appearance.}
    \label{fig:scenario}
\end{figure}

We deploy to rooms never seen during training (Table~\ref{tab:transfer}, Fig.~\ref{fig:scenario}). Without SEVO, ACT drops from 75\% in training to 30\% in novel similar environments and 0\% in extreme environments. SmolVLA follows the same pattern (70\%$\to$35\%$\to$0\%). With SEVO, success remains at 85\% (ACT) and 75\% (SmolVLA) in novel similar environments.

The extreme-environment drop (${\leq}$10\%) is an out-of-distribution issue: all training data was collected on dark-toned flooring, and the bright white laboratory floor exceeds the background variation seen during training. As noted in Section~\ref{sec:exp}, a separate training run using SEVO data collected in the laboratory confirms that this gap is closable by expanding the floor-tone diversity of the training set. Nonetheless, baseline policies achieve zero success in these settings, whereas SEVO still yields non-zero transfer.

\subsection{Data Efficiency}

ACT with SEVO achieves ${\sim}$80\% success with only 40 episodes, reaching 95\% at 80. SmolVLA fails to learn meaningful grasping below ${\sim}$80 episodes, reaching 83\% at 120. This higher data floor reflects the frozen-encoder bottleneck: the Action Expert needs more examples to learn to decode SEVO cues from compressed features, consistent with community recommendations of ${\geq}100$ episodes for SmolVLA~\cite{shukor2025smolvla}.

\subsection{Platform Generality}

Robot~B (IKEA LeRobot cart, Raspberry Pi~5, 8\,GB) with full SEVO achieves comparable success rates to Robot~A, despite a different camera layout (3 vs.\ 2 cameras), different physical design, and much slower inference. This consistency across platforms suggests that SEVO is not specific to Robot~A's hardware configuration.

\subsection{Effect of Wrist Cameras}

Wrist-mounted cameras are the default in most community setups: the original ACT/ALOHA system~\cite{zhao2023act} uses wrist cameras, SmolVLA's SO-100 evaluation~\cite{shukor2025smolvla} includes a wrist camera, and the LeRobot SO-ARM100 repository and LeKiwi mobile platform both provide official 3D-printable wrist camera mounts~\cite{soarm100_github}. Community members routinely add wrist cameras following these designs.

\begin{figure}[t!]
    \centering
    \includegraphics[width=\columnwidth]{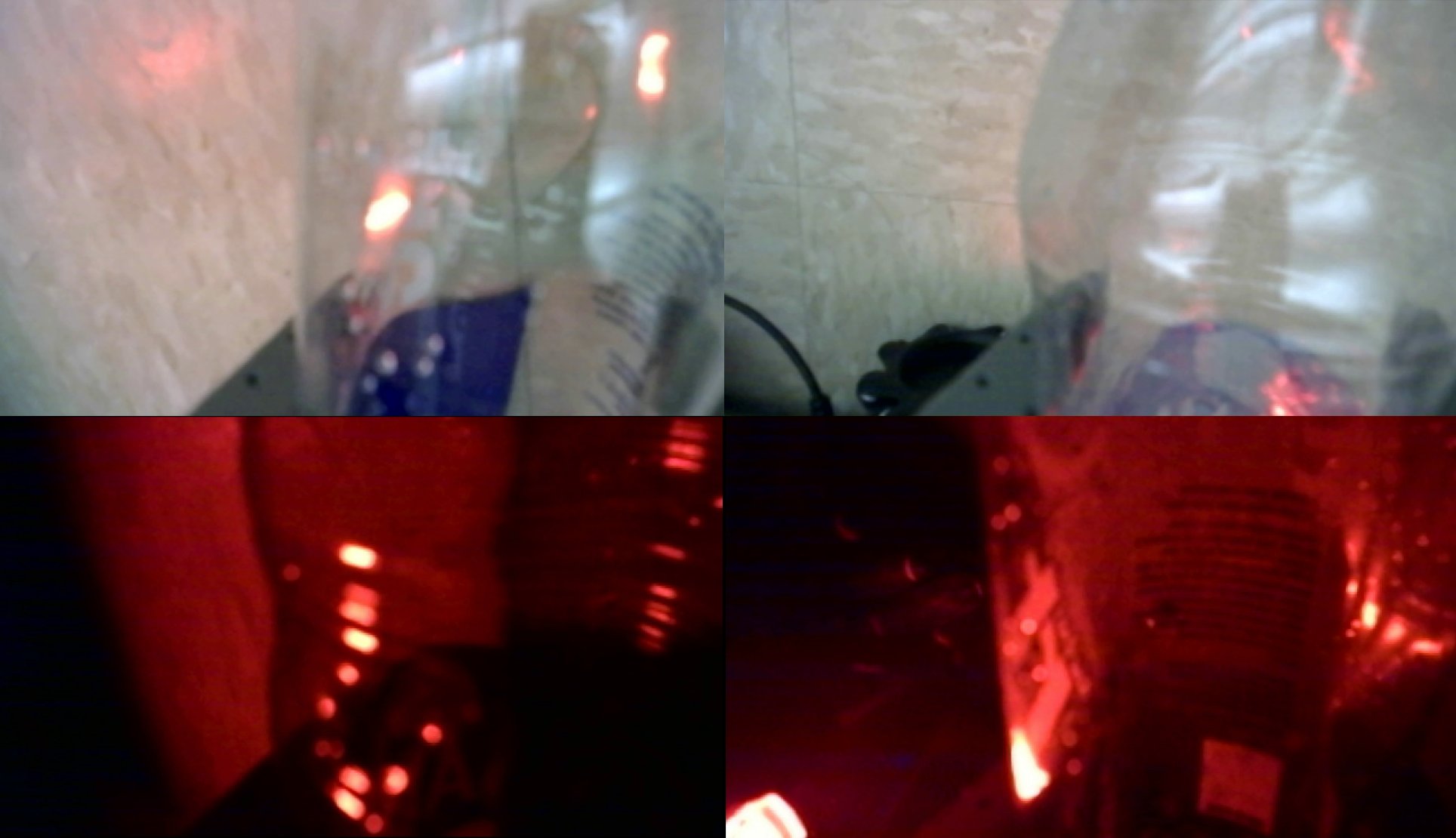}
    \caption{Wrist camera views during grasping on the SO-101 arm. The camera is mounted close to the gripper, resulting in a very limited field of view. \textbf{Top row:} ambient lighting; the bottle surface dominates the frame at close range. \textbf{Bottom row:} with red LED; the view is saturated by red reflections. In both cases, the wrist camera cannot convey gripper-to-bottle distance or global workspace context. During mobile operation, we observe oscillatory arm behavior with ACT and repeated grasp-release motions in open air with SmolVLA. By contrast, body-fixed cameras provide stable global views and the arm activates only upon confirmed bottle detection.}
    \label{fig:wrist}
\end{figure}

However, adding a wrist camera as a third input to Robot~A degrades performance considerably (Table~\ref{tab:ablation_wrist}, Fig.~\ref{fig:wrist}). We observe two distinct failure mechanisms:

\textbf{Spatial information poverty.} As the gripper approaches the target, the wrist camera view is increasingly dominated by the object surface, capturing only texture or red LED reflections with little spatial information about the gripper-to-object relationship. The YOLO mask fills the entire wrist frame at close range, further eliminating spatial cues. By contrast, body-fixed cameras maintain stable global views throughout the grasp sequence.

\textbf{Motion-induced failures.} During mobile operation, the wrist camera captures rapid visual flow. ACT exhibits continuous arm oscillation that fails to converge on the target; SmolVLA repeatedly executes grasp-release motions in open air without contacting the bottle ($\leq$12\% success). We hypothesize that ACT's fully trainable ResNet-18 backbone overfits to the dynamic wrist-camera patterns recorded during teleoperation, while SmolVLA's frozen SigLIP encoder maps rapid close-range views to ambiguous tokens that the Action Expert misinterprets as ``grasped'' states. Direct verification of these mechanisms is left to future work.

\begin{table}[t]
\caption{Component Ablation --- Wrist Camera Added (Robot~A)}
\label{tab:ablation_wrist}
\centering
\setlength{\tabcolsep}{3pt}
\begin{tabular}{ccccc}
\toprule
\textbf{YOLO} & \textbf{Red Light} & \textbf{Varied BG} & \textbf{ACT (\%)} & \textbf{SmolVLA (\%)} \\
\midrule
\cmark & \cmark & \cmark & 86 & 12 \\
\xmark & \cmark & \cmark & 76 & 3 \\
\cmark & \xmark & \cmark & 70 & 2 \\
\xmark & \xmark & \cmark & 62 & 1 \\
\xmark & \cmark & \xmark & 35 & 1 \\
\cmark & \cmark & \xmark & 22 & 1 \\
\cmark & \xmark & \xmark & 20 & 0 \\
\bottomrule
\end{tabular}

\vspace{1mm}
\begin{minipage}{\columnwidth}
\footnotesize{Same training data as Table~\ref{tab:ablation}, with wrist camera added as third input to Robot~A. $N{=}100$ trials per condition. ACT drops moderately (90\%$\to$86\%) while SmolVLA collapses (79\%$\to$12\%). Failure modes differ between the two policies; see Section~\ref{sec:exp}.}
\end{minipage}
\end{table}

\textbf{Static camera requirement.} These results show that SEVO requires cameras in fixed positions that observe the global workspace. The spatial information provided by wrist cameras (precise gripper-to-object distance) would be better served by a depth sensor than an RGB camera. Practitioners should prioritize \textit{stable observation geometry} over the intuitive appeal of an eye-in-hand view.

\subsection{Summary}

In summary, SEVO enables both ACT and SmolVLA to operate in cluttered, visually diverse environments with high success rates (Table~\ref{tab:ablation}, \ref{tab:transfer}). Cross-platform validation on Robot~B and wrist-camera ablations further support the generality of the approach and the importance of static camera placement.

\section{DISCUSSION}

\subsection{Why SEVO Benefits ACT More Than SmolVLA}

We hypothesize that the 12-point gap (95\% vs.\ 83\%) follows from the architectural difference detailed in Section~\ref{sec:architecture}. ACT's trainable \texttt{conv1} adapts directly to SEVO's color signatures, and deeper layers build grasp-relevant representations end-to-end. SmolVLA's frozen SigLIP compresses the same signals through fixed parameters, and the trainable Action Expert must decode them indirectly. Diversified backgrounds amplify this gap because background variation acts as a regularizer at the gradient level, an effect that is stronger when gradients reach the visual encoder directly (ACT) than when they only update a downstream module (SmolVLA).

\subsection{Limitations and Future Work}

\textbf{Static grasping only.} The mobile base must stop for ${\sim}$2\,s before grasp execution. Grasping while moving drops success to ${\sim}$10\%.

\textbf{Extreme environment gap.} SEVO's robustness is bounded by the visual diversity of the training distribution. All training data in our primary experiments was collected on dark-toned floors; deployment on a bright white laboratory floor caused sharp degradation (${\leq}$10\%). A separate training run conducted entirely in the white-floor lab confirmed that SEVO works in that setting, indicating that broader floor-tone coverage during collection would likely eliminate this gap.

\textbf{YOLO dependency.} Performance depends on YOLO detection quality; novel categories or heavy occlusion degrade the overlay. Open-vocabulary detectors could replace YOLO.

\textbf{SmolVLA latency.} SmolVLA requires ${\sim}$2\,min per grasp on Jetson Orin NX and ${\sim}$15\,min on Raspberry Pi~5, limiting practical deployment. ACT runs in real time on both platforms.

\section{CONCLUSION}

We presented SEVO, a data-centric method for robust real-world manipulation using standard RGB cameras and community-accessible hardware. Through systematic ablation on two mobile platforms with both ACT and SmolVLA, we found that diversified data collection is the single most important factor for cross-environment generalization, outranking active illumination and YOLO overlay. Body-fixed cameras substantially outperform wrist cameras, which induce severe policy-dependent failures. The full SEVO pipeline achieves 95\% (ACT) and 83\% (SmolVLA) in training, transferring to novel environments at 85\% and 75\%, whereas the same policies without SEVO drop to 30\% and 35\%. Parameter-level inspection shows that SEVO's observation-space modifications are most effective when the visual encoder is fully trainable (ACT) and less so when frozen (SmolVLA). Taken together, these results suggest that principled observation design and environmental diversity during data collection can substitute for model scaling on low-cost hardware.

\addtolength{\textheight}{-1cm}

\bibliographystyle{IEEEtran}
\bibliography{refs}

\end{document}